\documentclass{comjnl}
\usepackage{amsmath}
\usepackage{graphicx}
\usepackage{grffile}
\usepackage{xspace}
\DeclareUnicodeCharacter{2212}{\textendash}

\begin{document}

\title[Estimation of Warfarin Dosage with Reinforcement
Learning]{Estimation of Warfarin Dosage with Reinforcement
Learning}
\author{Arpita Vats}
\affiliation{Department of Computer Science, Boston University, Boston,USA} \email{arpita@bu.edu}

\shortauthors{Arpita Vats}
 

\begin{abstract}
In this paper it has attempted to use Reinforcement learning to model the proper dosage of Warfarin for patients.The paper first examines two baselines: a fixed model of 35 mg/week dosages and a linear model that relies on patient data. We implemented a LinUCB bandit that improved performance measured on regret and percent incorrect. On top of the LinUCB bandit we experimented with online supervised learning and reward reshaping to boost performance. Our results clearly beat the baselines and show promise of using multi-armed bandits and artificial intelligence to aid physicians in deciding proper dosages.
\end{abstract}

\maketitle

\section{Introduction}
Warfarin is the most widely used oral blood anticoagulant agent worldwide; with more than 30 million prescriptions for the drug in the United States in 2004.The correct dose of warfarin is difficult to establish because it vary subsequently for every patients, and the consequences of taking incorrect dosage is severe. If a patient receives a dosage that is too high, they may experience inadequate anticoagulation
(which means that it is not helping prevent blood clots). Since incorrect doses contribute to a high rate of adverse effects, pursuing improved strategies for determining an appropriate dose is highly desired.(International Warfarin Pharmacogenetics Consortium, 2009).\\
 In practice a patient is typically prescribed an initial dose, the doctor then monitors how the patient responds to the dosage and then adjusts the patient’s dosage as treatment progresses. This interaction can proceed for several rounds before the best dosage is identified. However it is best if the correct dosage can be initially prescribed to prevent the adverse effects described earlier. The project is motivated by the challenge of Warfarin dosing, and considers a simplification of this important problem, using real data. The goal of this project is to explore the performance of multi-armed bandit algorithms and artificial intelligence to best predict the correct dosage of Warfarin for patients without a trial -an-error procedure as typically employed.\\
 We use publicly available patient dataset that was collected by staff at the Pharmacogenetics and Pharmacogenomics knowledge base(PharmGKB) for 5700 patients who were treated with Warfarin from 21 research groups spanning 9 countries and 4 continents. Features of each patient in this dataset includes, demographic (gender, race, ….), back-gorund(height, weight, medical history,...), phenotypes and genotypes. There are in total 5,528 patients with the known therapeutic dose of Warfarin in the dataset. Given this data one can classify the right doses for each patient as low : less than 21mg/week, medium : 21-49 mg/week, and high : more than 49 mg/week.

\section{Background and Related Work}

\subsection{Problem Background}
Warfarin is a prescription medication used to prevent harmful blood clots from forming or growing larger. Giving patients the proper dose is very challenging as dose varies from patient due to genetic variability. The incorrect dosage can contribute to many adverse effects and may hold serious consequences to the health of the patient. The importance of getting a correct dose of Warfarin is imperative and this led to our desire to construct a dose of warfarin for patients. The current method is to give patients a small dose and slowly increment until the desired results occur but this is a slow and potentially dangerous method.

\subsection{Related Work} 
Given the importance of the problem, there has been previous work addressing how to best model Warfarin dosing. Two standouts paper have addressed the problem are the The International Warfarin Pharmacogenetics Consortium’s publication of “Estimation of the Warfarin Dose with Clinical and Pharmacogenetic Data” and Bastani and Bayati’s work “Online Decision-Making with High-Dimensional Covariates”.
\subsection{Estimation of the Warfarin Dose with Clinical and Pharmacogenetic Data}
The International Warfarin Pharmacogenetics Consortium published their work “Estimation of the Warfarin Dose with Clinical and Pharmacogenetic Data '' in 2009. This paper curated the PharmGKB dataset mentioned previously and focused on clinical factors, demographic variables, and variations in two genes — cytochrome P450, family 2, subfamily C, polypeptide 9 (CYP2C9), and vitamin K epoxide reductase complex, subunit 1 (VKORC1) (International Warfarin Pharmacogenetics Consortium, 2009). \\The consortium found that the best performing model, performance gauged on the criterion of lowest mean absolute error, was an ordinary least-squares linear regression that predicts the square root of the dose and incorporates both genetic and clinical data. This pharmacogenetic model outperformed the clinical model as well as the fixed dosed model substantially. A key takeaway was that patients who most benefited from the pharmacogenetic model are on the extremes of the Warfarin spectrum such that an underdose or overdose will cause significant harm. This is something we explore further with reward reshaping.

\subsection{ Online Decision-Making with High-Dimensional Covariates}
This paper focuses on the idea of personalizing decisions based on individual-level conditions or requirements. The paper attempts to address the problem of learning a model of decision rewards conditional on individual specific covariates and presents the application of personalized medicines and in particular mentions the Warfarin problem.The paper makes also calls to The International Warfarin Pharmacogenetics Consortium’s work on the Warfarin problem and uses the same PharmGKB dataset. This paper uses a LASSO Bandit and this bandit outperforms fixed dosing by physicians. Despite the assumption that every patient dose can be examined as a bandit problem, the LASSO Bandit showed that reinforcement learning can be effective in improving performance on tasks such as the Warfarin problem where personalized decisions on the individual-level are beneficial.

\section{ Approach	}
\subsection{ Data Preprocessing}
The data set we used was the PharmGKB data set curated by the International Warfarin Pharmacogenetics Consortium. The dataset consists of information about each patient, including gender, race, ethnicity, age, height, and so on, as well as the correct Warfarin dosage. We dropped 173 patients without information on correct Warfarin dosage.The dataset still had other key information missing for some patients such as the Values for age, height and weight. To fill these values in, we took the mode of the age to get an approximate age in decades and the mean of heights and weights in the data set. Other fields that potentially had missing values, we decided to treat a missing value as a possible value the feature could take on. This left our dataset with 5,528 valid entries.

\subsection{Overview of Approaches}
To address the Warfarin dosage prediction problem, we used a variety of methods to gauge how to model Warfarin dosing. As the first baseline, we used a fixed dose model. The second baseline was a linear model. Our third model was a LinUCB Bandit. The fourth model was online Supervised Learning. The last model was reshaping the reward structure for both the Bandit and Supervised Learning.

\subsection{Baseline 1: Fixed Dose}
This model was a fixed dosed model that would give up all patient doses of 35 mg/week. This approach was important to gauge how assigning all patients a conservative medium dose would perform. Due to the dataset having the correct dose of warfarin for all patients (Therapeutic Dose of Warfarin column), it was simple to gauge if a medium dose was the adequate amount.

\subsection{ Baseline 2: Linear Model}
For the second baseline, the model we used was the Warfarin Clinical Dosing Algorithm. This model was a linear model that had predetermined weights for the features required. The features required for this clinical dosing linear model were Age in decades, Height in cm, Weight in kg, if patient is Asian, if the patient is Black or African American, if patient is of mixed or missing race, if patient is taking one of Carbamazepine, Phenytoin, Rifampin, or Rifampicin and lastly if the patient is taking Amiodarone. This linear combination of features would then give us the square root of the weekly dose.\\
\graphicspath{{/Images/Image2.png}}
\includegraphics[width = 7cm]{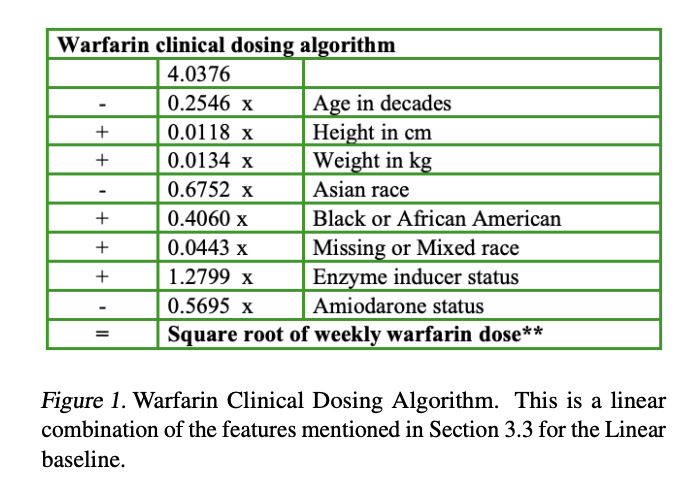} \\

\subsection{Feature Engineering}
Before going into a more robust model, we wanted to perform some feature engineering and find that feature would give us better performance and were more important to the Warfarin Problem. After many iterations we settled upon 26 features to represent each patient. These feature are :
\begin{itemize}
\item 
Age in decades
\item
Height in cm
\item
Weight in kg
\item
Race (Indicators of isAsian, isBlack, isMissing)
\item
Whether the patient is taking Amiodarone
\item
Gender
\item
VKORC1 genotype
\item
VKORC1 QC genotype
\item
Enzyme inducer status(whether the patient is taking Carbamazepine, Phenytoin, Rifampin or Rifampicin)

\end{itemize}
For categorical features, we use the standard one-hot vector encoding.

\subsection{ LinUCB Bandit}
LinUCB (Li et al., 2010) is an upper confidence bound algorithm that assumes that the reward is a linear function of the d features of a context.That is, for reward $\/r_{t,a}$ Context $\/x_t$  arm a, and weight vector $\hat{\theta}_a $  we assume
$E[r_{t,a} \mid x_t] = x_t ^T\hat{\theta}_a $ .For every patient, the algorithm will perform least squares linear regression with all previously seen data and construct a model to predict reward for each arm. Then,the algorithm simply chooses the arm that gives the expected payoff. A more detailed explanation can be found in Algorithm 1. Our choice  of $\alpha$ was initially chosen to be 1.0. \\

\textbf{Algorithm 1  :- LinUCB for Contextual Bandits}

\begin{enumerate}
    \item \textbf{for} each arm \textbf{do}
    \item $A_a \leftarrow I_d$
    \item $b_a \leftarrow 0$
    \item end for
    \item \textbf{for} t = 1......N \textbf{do}
    \item observe context $X_t$
    \item \textbf{for} each arm a \textbf{do}
    \item $ \hat{\theta} \leftarrow A_a ^{- 1} b_a$
    \item $ p_{t,a} \leftarrow x_t^T\hat{\theta}_a + \sqrt{x_t^TA_a^{-1}x_t}$
    \item \textbf{end for}
    \item choose action  $a_t \leftarrow \underset{a}{\arg \max}$ $ p_{t,a}$
    \item observe reward $ r_t$
    \item $A_{a t} \leftarrow A_{a,t} + x_t x_t^T$
    \item $b_{a t}\leftarrow b_{a t}+ r_t x_t$
    \item \textbf{end for}
\end{enumerate}

The LinUCB algorithm , like other upper confidence bound algorithms, admits  sublinear regret  $  \Tilde{\mathcal{O}} \sqrt{KdT}  $ for K arms, d features, and T steps. The  notation indicates that we ignore logarithmic terms.

\subsection{Online Supervised Learning}
We additionally examine supervised  learning in an online (i.e. bandit like) setting. For every patient, we fit a model that predicts correct dosage from contexts. This model is necessarily stronger than that used in LinUCB, as LinUCB performs regression on rewards as opposed to correct dosages. Regression directly on the correct dosages provides finer-grained information compared to the reward observed, allowing our supervised learning approach to dominate LinUCB. An outline of the algorithm is shown in Algorithm 2. \\

{\textbf{Algorithm 2 : Online Supervised Learning for Contextual Bandits }} \\

\begin{enumerate}
    \item \textbf{for} t = 1....N \textbf{do}
    \item observe context $x_t$
    \item $ \textbf{A} \leftarrow$ matrix of all previously seen context 
    \item $\textbf{b} \leftarrow $ vector of correct dosage of all previously seen contexts
    \item $\theta \leftarrow A^{- 1}b$
    \item choose action dictated by $ x_t ^T \theta $
    \item observe reward $r_t$
    \item \textbf{end for}
\end{enumerate}

While the algorithm shown in Algorithm 2 specifically shows linear regression for modeling correct dosages,stronger function approximators may be used. Neural networks in particular show great promise for modeling complex, nonlinear functions. The difficulty in using neural networks in the online supervised learning paradigm is with optimization – rather than simply computing a pseudo inverse in regression, the parameters of the neural network must be trained through gradient descent or some other optimization method. These methods are often more computationally intensive and sometimes intractable. Different heuristics may be applied to alleviate this problem, such as only retraining every k steps. However, the online supervised learning problem continues to be difficult and further work must be done for this to be a tractable solution.
\subsection{Reward Reshaping}
Following the work of The International Warfarin Pharmacogenetics Consortium, we were inspired to pursue reward reshaping due to how giving a patient an underdose or overdose can produce significant harm. Since the goal of Warfarin is to prevent clots and serve as a blood thinner, we believed that giving a patient who needed a high does a low dose was worse than giving a patient who needed a low dose a high dose as the patient would still be at risk to a clot if they needed a high does and got a low one. Further research into the effects of Warfarin inspired us to reshape our reward structure. If we gave the patient the correct dose, then the reward of 0 was appropriate. Giving a patient who needed a low dose a high dose was going to be −R where R = 1.5. Giving a high dose patient a low dose would be −2 × R. Every other scenario was decided to be −R 2 . The final reward reshaping table can be seen in Figure 2.
\par
\graphicspath{{/Images/Image2.png}}
\includegraphics[width = 7cm]{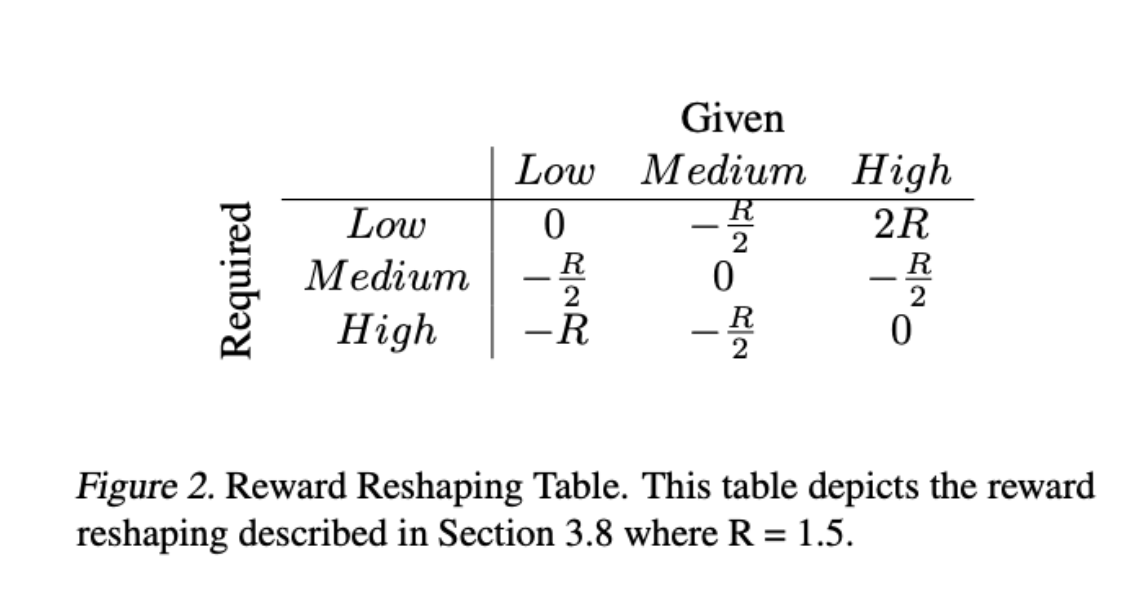}
\section{Experiment results }
\subsection{Evaluation Metrics}
We use two evaluation metrics: Regret and Fraction of incorrect dosing decisions.\\
To calculate Regret, we model the reward for each arm $a_i$ for a patient with feature $X_t$ as: 
\par
\begin{equation}
\label{eq1}
\begin{split}
   r_t(X_t, a_i) = X_t^T\beta_i + \epsilon_{i,t}
\end{split}
\end{equation}

and find $\beta_i$ for each arm by calculating the least square area solution to the system of 5,528 linear equations (one for each patient) where $X_t$ is the 26 features we have previously described. This works for our Linear Model baseline as well.Since the 26 features include all features used in the Linear Model baseline.\\
For our reward reshaping experiments, we refit $\beta_i$ for each arm since $r_t(X_t, a_i)$ has been changed as previously described in Figure 2.\\ Under this model, if the agent chooses arm i for patient t, it
will incur the expected regret of:
\begin{equation}
\label{eq1}
\begin{split}
  [\underset{j}{max}[X_t^T \beta_j]- X_t^T\beta_i]
\end{split}
\end{equation}
The calculation of Fraction of incorrect dosing decisions,which is the fraction of incorrect dosing decisions made so far by the algorithm at the current timestep out of all patients seen so far by the algorithm at the current timestep,is straightforward. We additionally report the Fraction of right decisions for the two baseline methods as requested.
\subsection{Baseline}
\subsubsection{Baseline 1: Fixed Dosage}
Figure 3 shows the fraction of right decisions that the fixed dose baseline makes. We notice that the fraction of right decisions fluctuates initially when there are few patients seen, and eventually settles at 0.6118.\\
\graphicspath{{/Images/Image3.png}}
\includegraphics[width = 7cm]{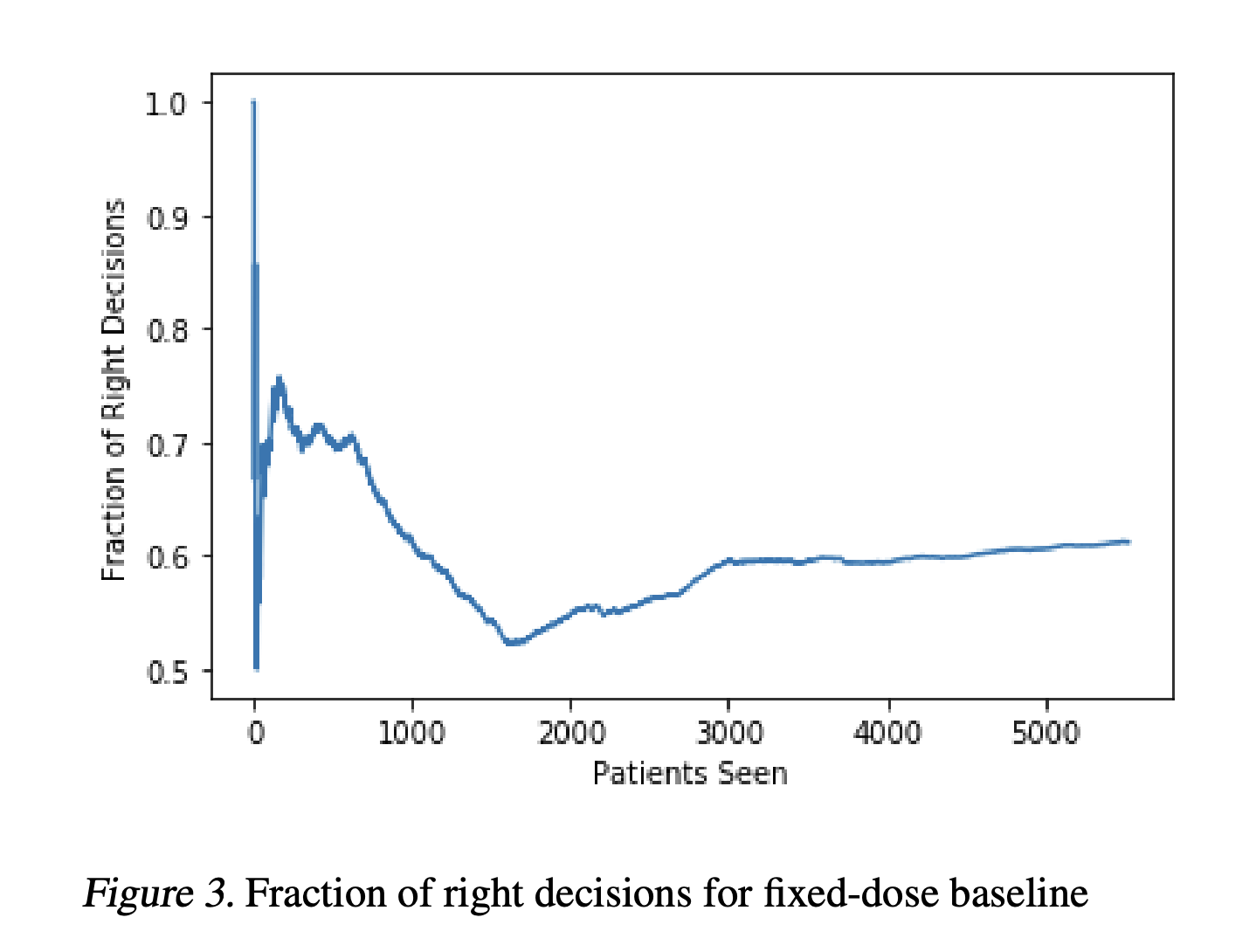}\\
\subsubsection{Baseline 2 : Linear Model}
Figure 4 shows the fraction of right decisions that the Linear Model baseline makes. We notice that similar to the fixed-dose baseline, the fraction of right decisions fluctuates initially when there are few patients seen, but the fluctuation is less. The final fraction of right decisions after all patients are seen is 0.6431, which is higher than that of the fixed dose baseline. From Figure 6, we cans see more clearly that linear model baseline, as expected, clearly outperforms fixed-dose baseline by having a lower fraction of incorrect decisions. Figure 5 also shows how linear model baseline has much lower regret compared to fixed-dose baseline as expected.\\
\graphicspath{{/Images/Image4.png}}
\includegraphics[width = 7cm]{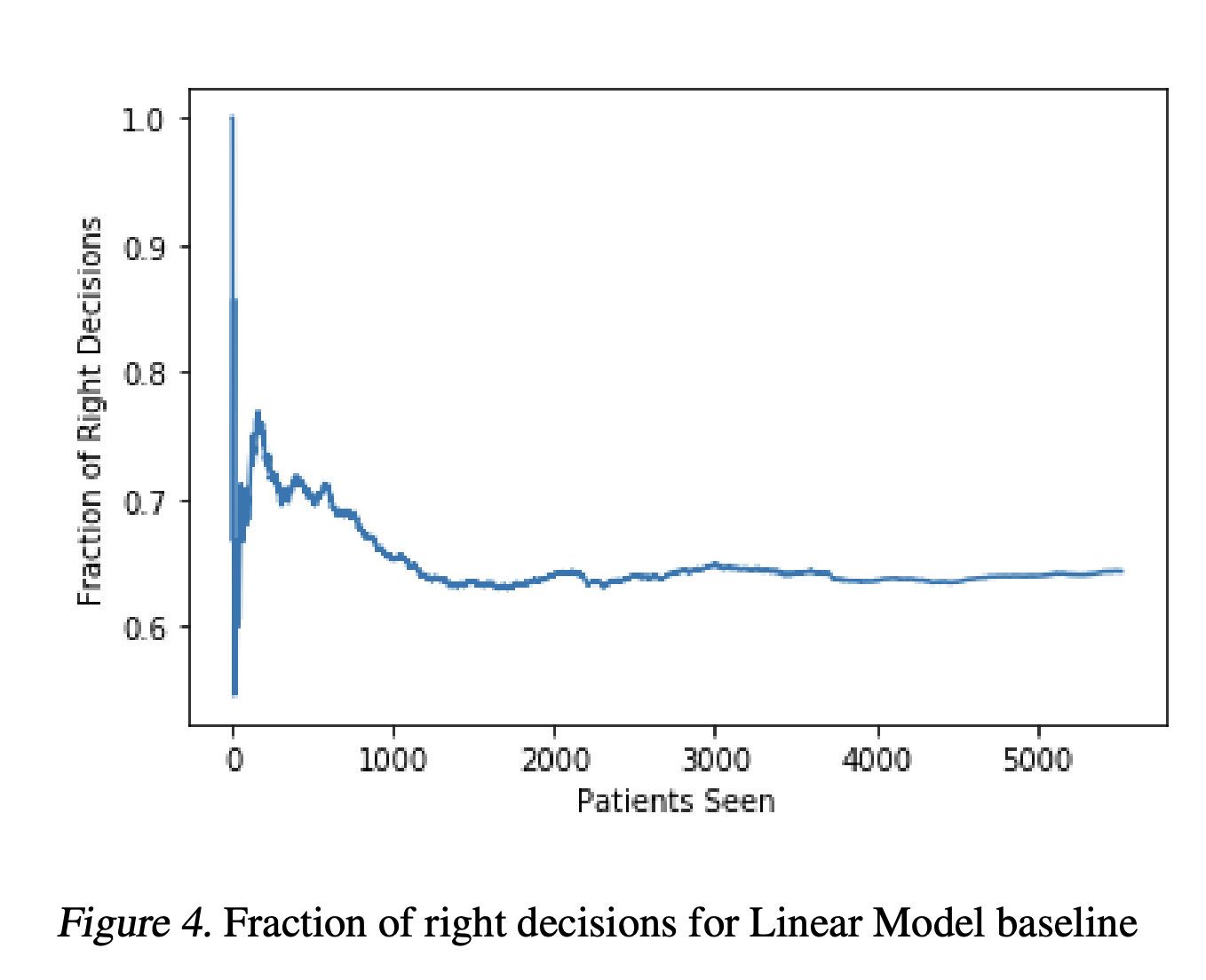}\\
\subsection{LinUCB}
After running the LinUCB model 20 times on different shuffling of the dataset and finding the $95\%$ confidence interval (using the T-Distribution) for both the cumulative regret and incorrect fraction, we plotted the results for the LinUCB against the results from baseline 1, the fixed dosed model, and baseline 2, the linear model.\\
\graphicspath{{/Images/Image5.png}}
\includegraphics[width = 7cm]{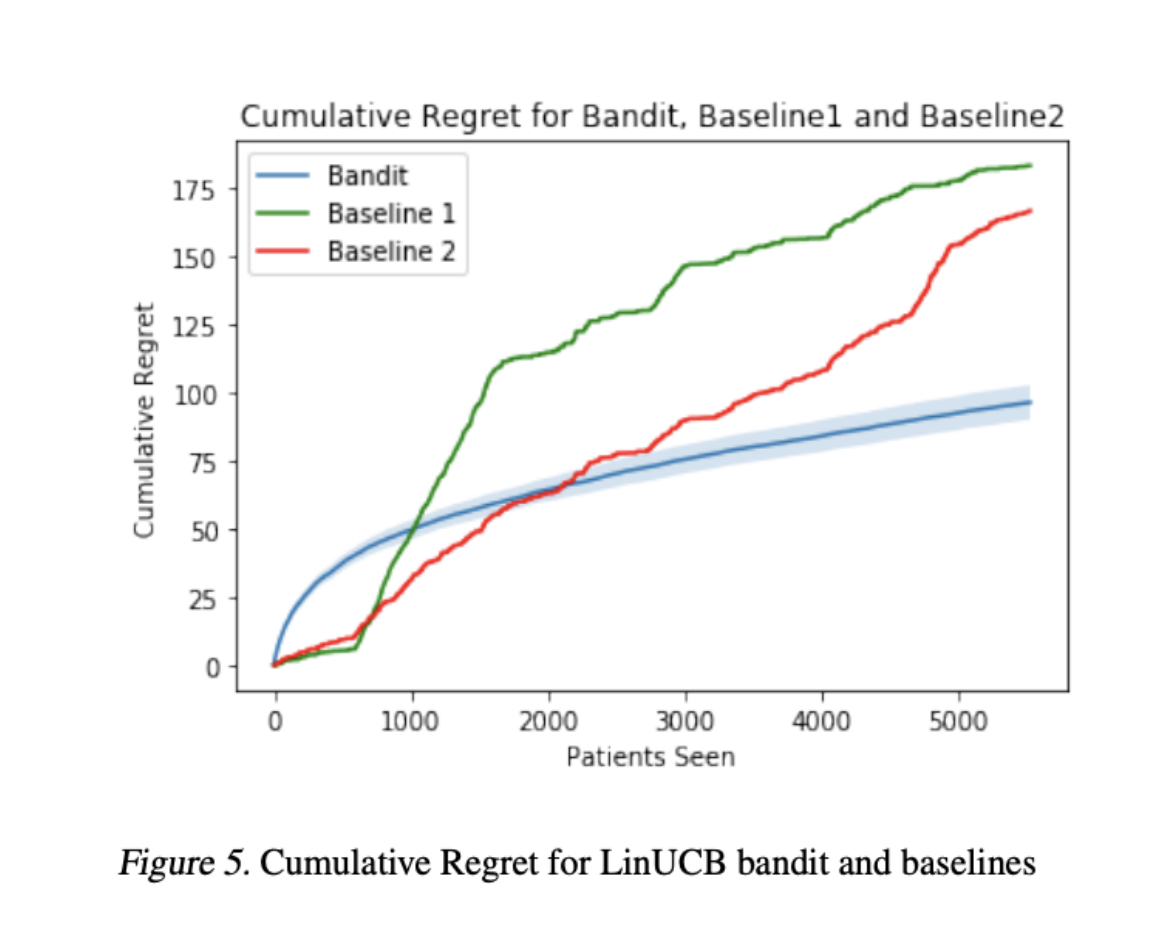}\\
In Figure 5, we see that the cumulative regret for baseline 2, the linear model, is lower than that of baseline 1, fixed dose. The key takeaway however is that we clearly see that the regret for the bandit is sublinear, which indicates that as we continue to see more patients, the LinUCB bandit will maintain a lower regret than the baseline. Based on cumulative regret, using the LinUCB model is promising.\\
\graphicspath{{/Images/Image6.png}}
\includegraphics[width = 7cm]{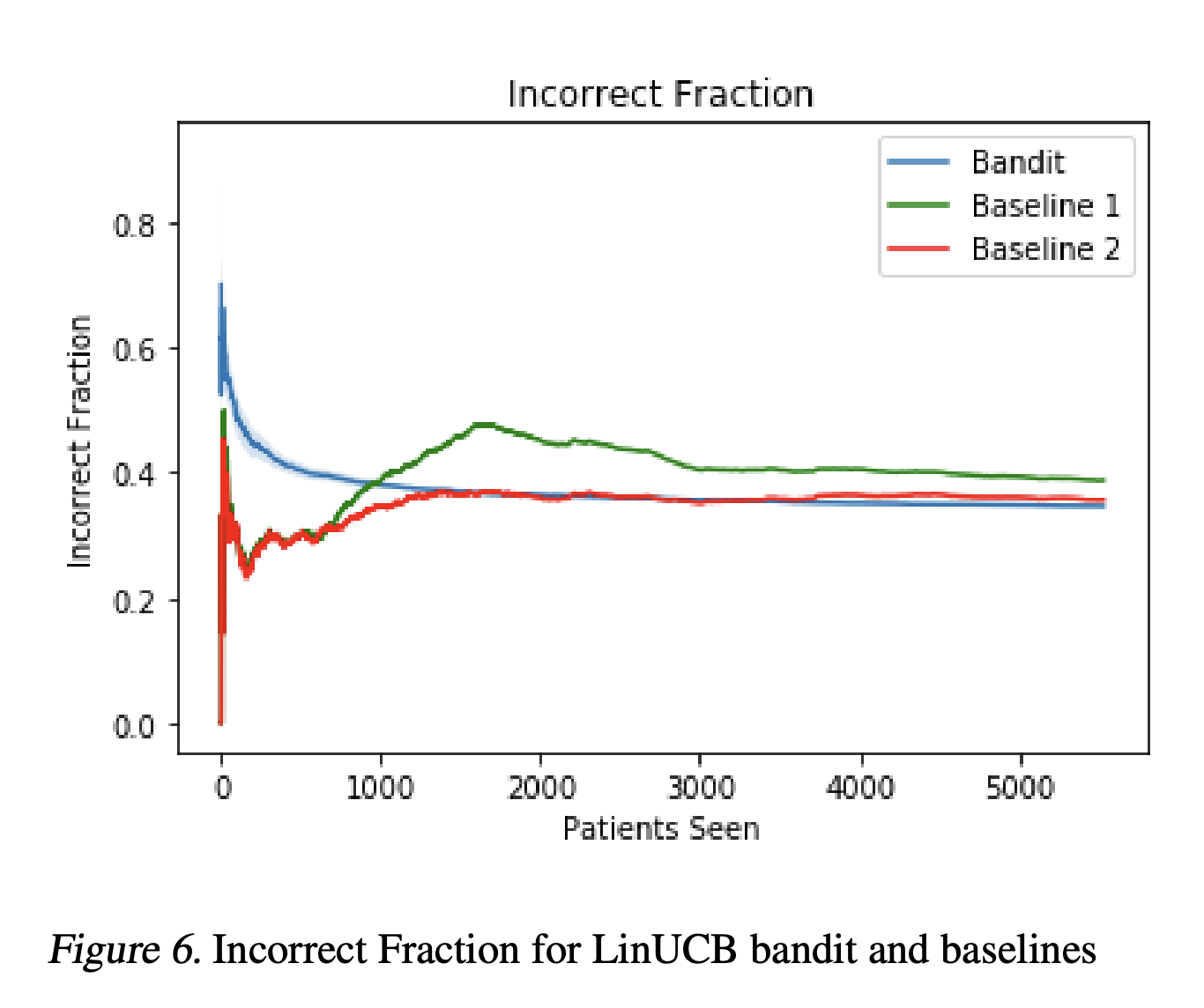}\\
Figure 6 shows the performance of the LinUCB bandit versus the two other baselines measuring incorrect fraction. It is difficult to see in Figure 6 that the LinUCB model beats the baselines by achieving a lower incorrect fraction but in Figure 7, the zoomed-in version, it is clear that LinUCB is a better fit model for the Warfarin Problem.\\
\graphicspath{{/Images/Image7.png}}
\includegraphics[width = 7cm]{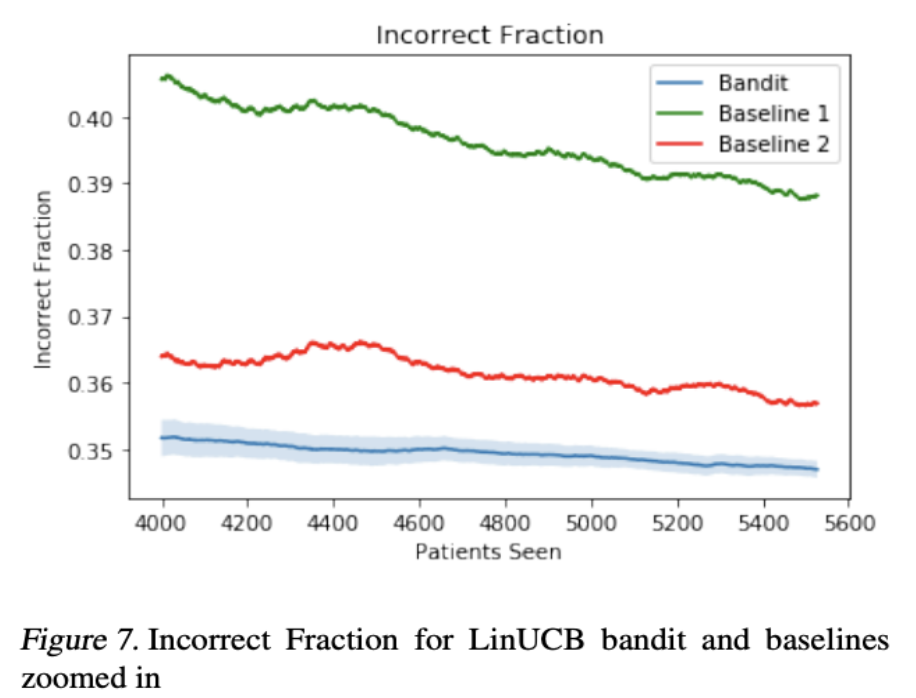}\\
\subsection{Performance of Online Supervised Learning}
Keeping the same metrics of performance, cumulative regret and incorrect fraction, we had our online supervised learning model go up against the two baselines and the LinUCB Bandit.\\
\graphicspath{{/Images/Image8.png}}
\includegraphics[width = 7cm]{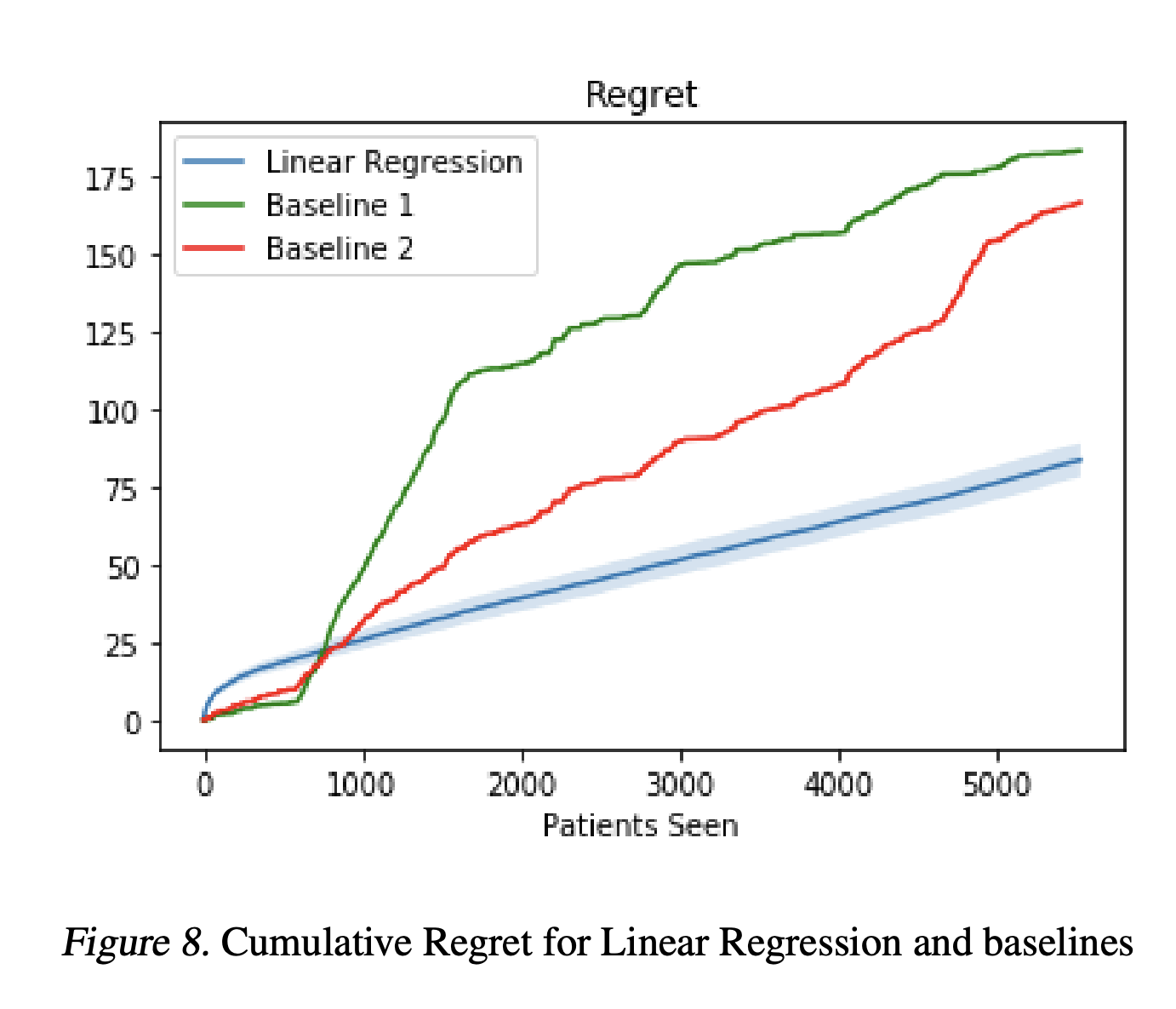}\\
\graphicspath{{/Images/Image9.png}}
\includegraphics[width = 7cm]{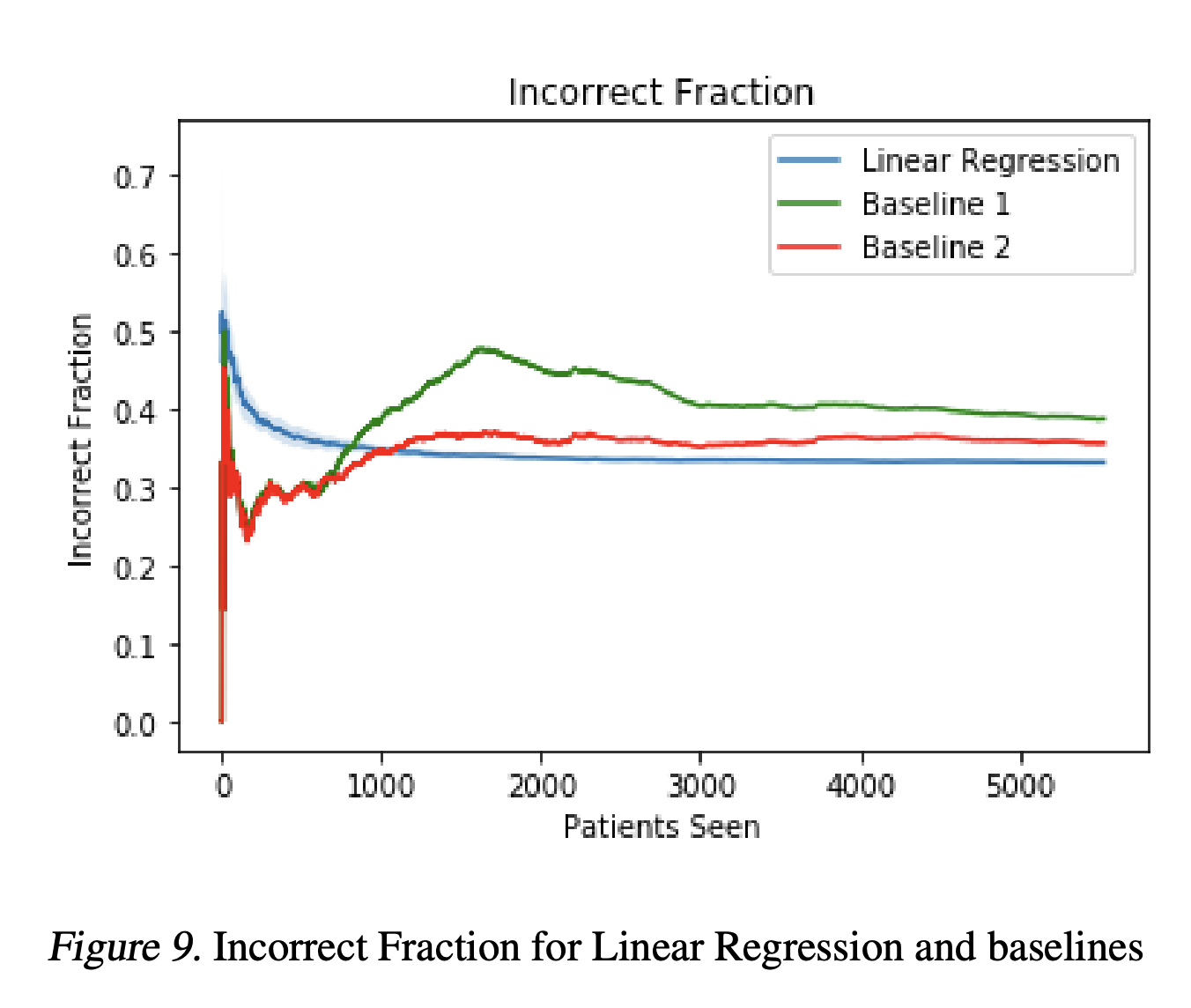}\\
From Figure 8 and Figure 9, we see that the Linear Regression clearly surpassed the baselines using the two metrics of performance. Unlike the LinUCB bandit, we see that the Linear Regression model clearly beats the baselines using the metric of Incorrect Fraction. To see if the Linear Regression model is a good fit for the Warfarin problem, we must compare the results to the LinUCB model. In Figure 10, we see that for the number of patients that we saw, Linear Regression had a lower cumulative regret than that of the LinUCB Bandit. After 5,528 patients, the Linear Regression is lower than the bandit and does not overlap in confidence interval but the behavior of their regrets call into question whether the Linear Regression will be able sustain the sub-linear regret the LinUCB model demonstrates.\\
\graphicspath{{/Images/Image10.png}}
\includegraphics[width = 7cm]{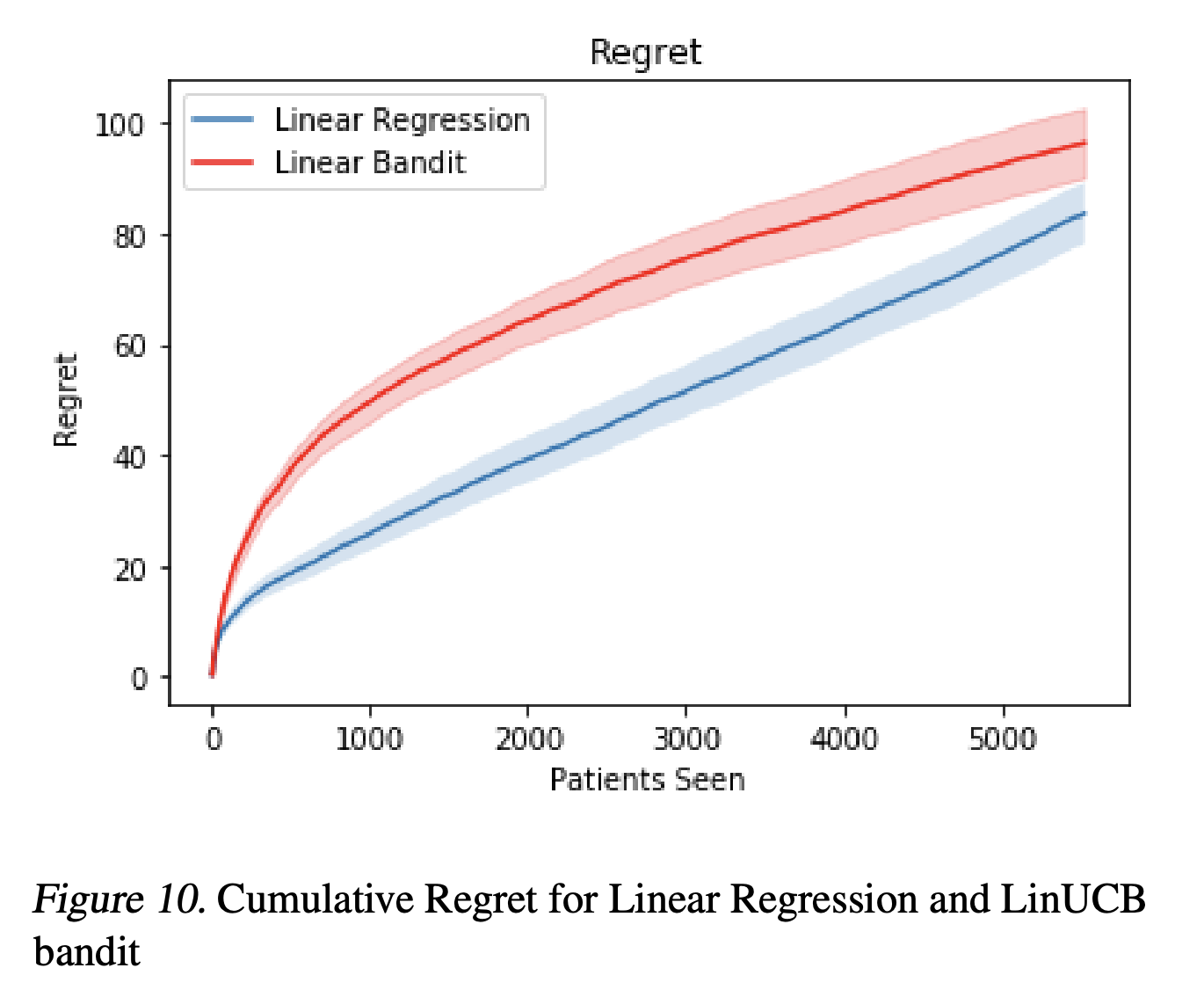}\\
\graphicspath{{/Images/Image11.png}}
\includegraphics[width = 7cm]{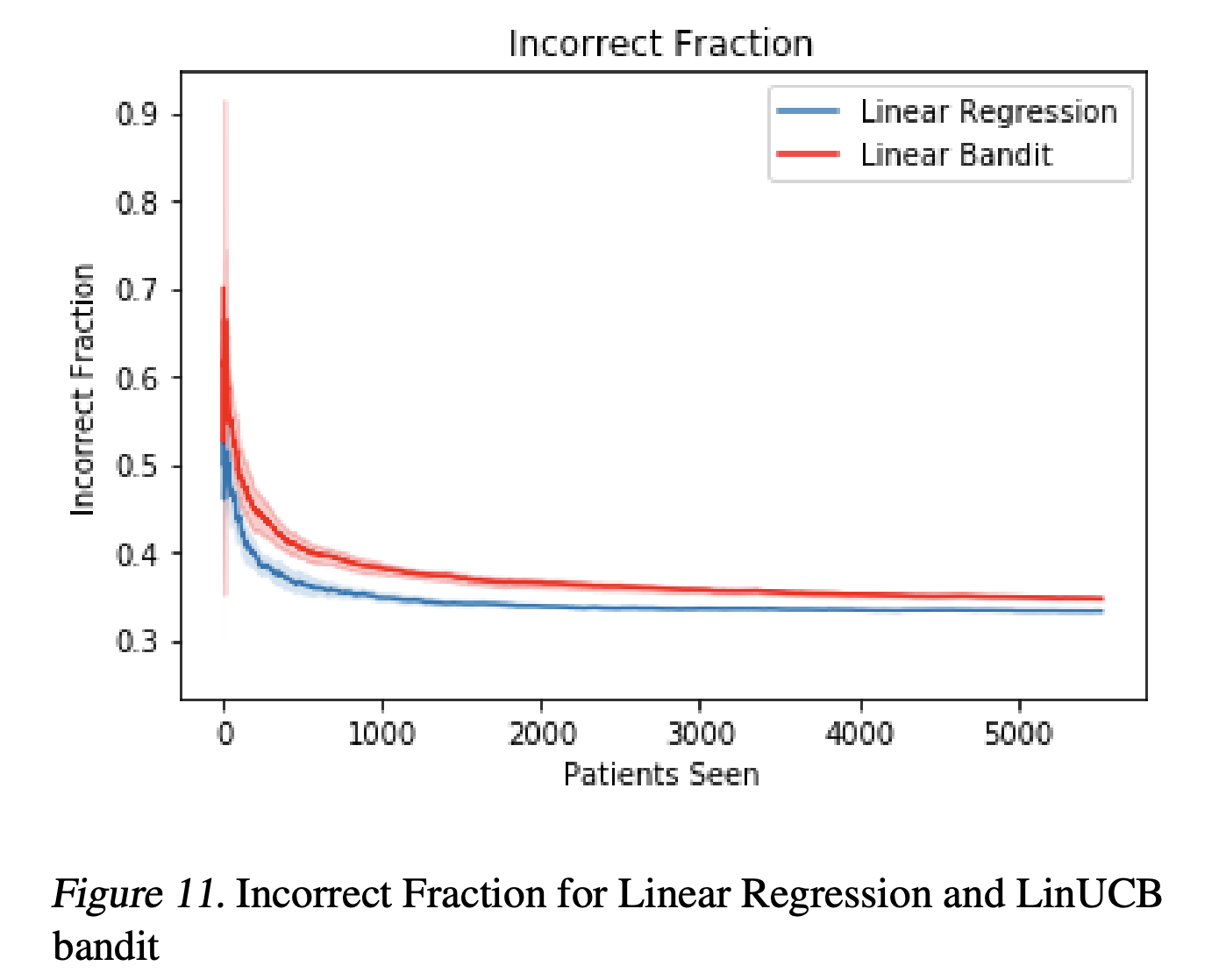}\\
However, from figure 11, we can see that the linear regression model clearly outperforms the LinUCB bandit model on incorrect fraction, as seen by how the blue line, representing the incorrect prediction fraction for linear regression, clearly lies below the red line, representing incorrect prediction fraction for LinUCB, with no overlap in confidence interval after the 1000 patients. This is a clear sign of success for the supervised learning model and is indicative that the model can be well equipped to handle the Warfarin dosage prediction problem.
\subsection{Reward Reshaping Reward}
Reshaping was an idea that we felt would be very successful in addressing this problem due to the importance of getting the proper dosage and avoiding potential adverse effects of getting an incorrect dose. Reward reshaping directly impacts how regret is computed so to make the comparisons fair, we applied reward reshaping to all of our models for comparison.\\
\graphicspath{{/Images/Image12.png}}
\includegraphics[width = 7cm]{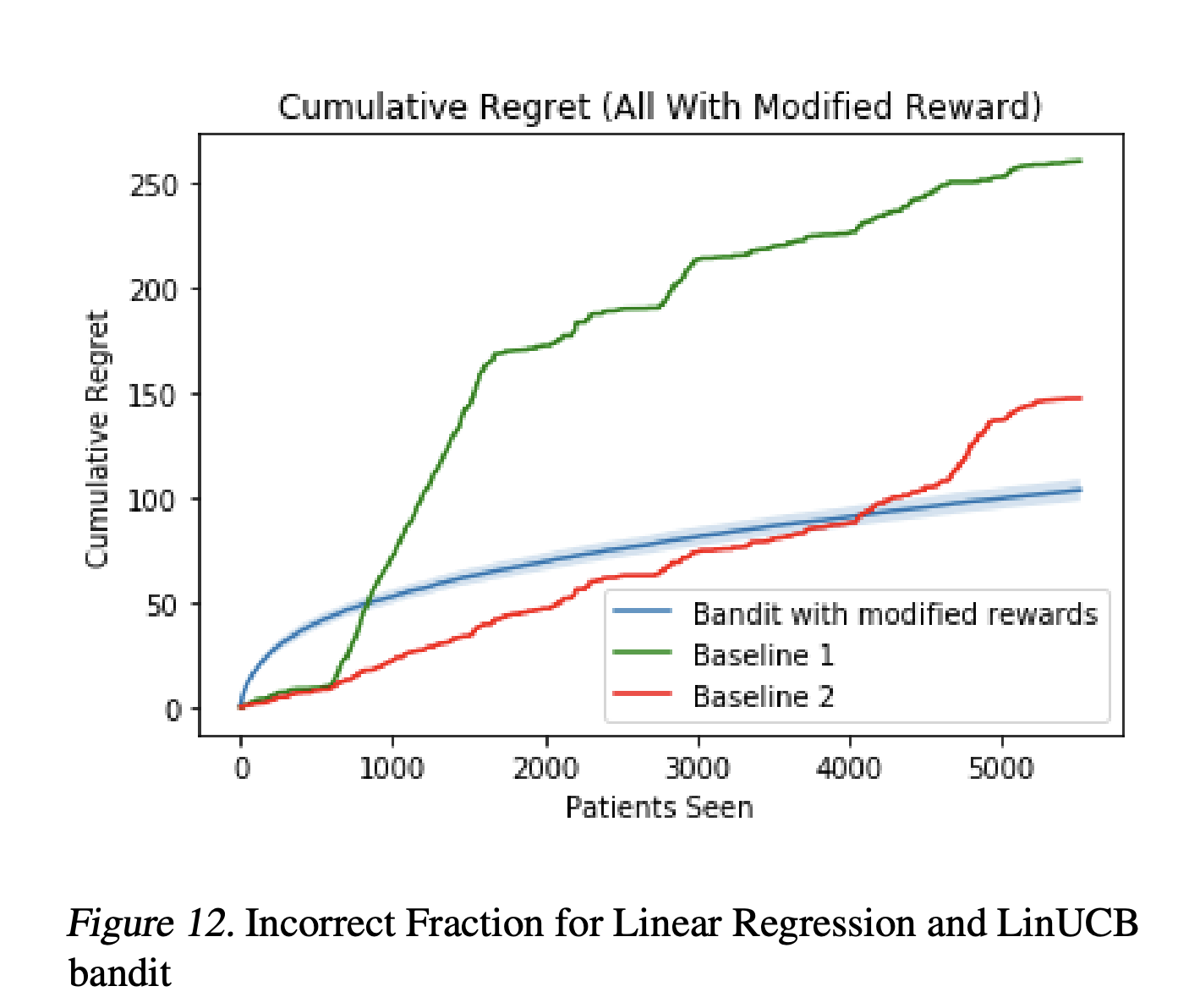}\\
In Figure 12, we see that the reward reshaping alters the cumulative regret of all the models when compared to Figure 5. The key takeaway here is that not only did reward reshaping decrease the cumulative regret for the LinUCB bandit compared to that in Figure 5 but it also very significantly decreased the cumulative regret of the Linear baseline. This result begins to indicate that reward reshaping can be effective for certain problems. Figure 13 depicts how the reward reshaping still allows for the LinUCB model to have a lower incorrect fraction than the two baselines. This solidifies that LinUCB is a better model than the two baselines as its performance was consistent despite changes to the reward structure. When comparing the performance of the LinUCB bandit with the standard reward structure to that of LinUCB with reward reshaping (Figure 14), we can see that the model with reward reshaping generally performed better than the one with standard rewards as seen in its lower fraction of incorrect decision. The performance gap, however, is very small as both models are in each other’s confidence intervals. However, the results are still indicative that there are potential benefits in altering the reward structure of the models. More work could be done to find the best reward structure for the problem.\\
\graphicspath{{/Images/Image13.png}}
\includegraphics[width = 7cm]{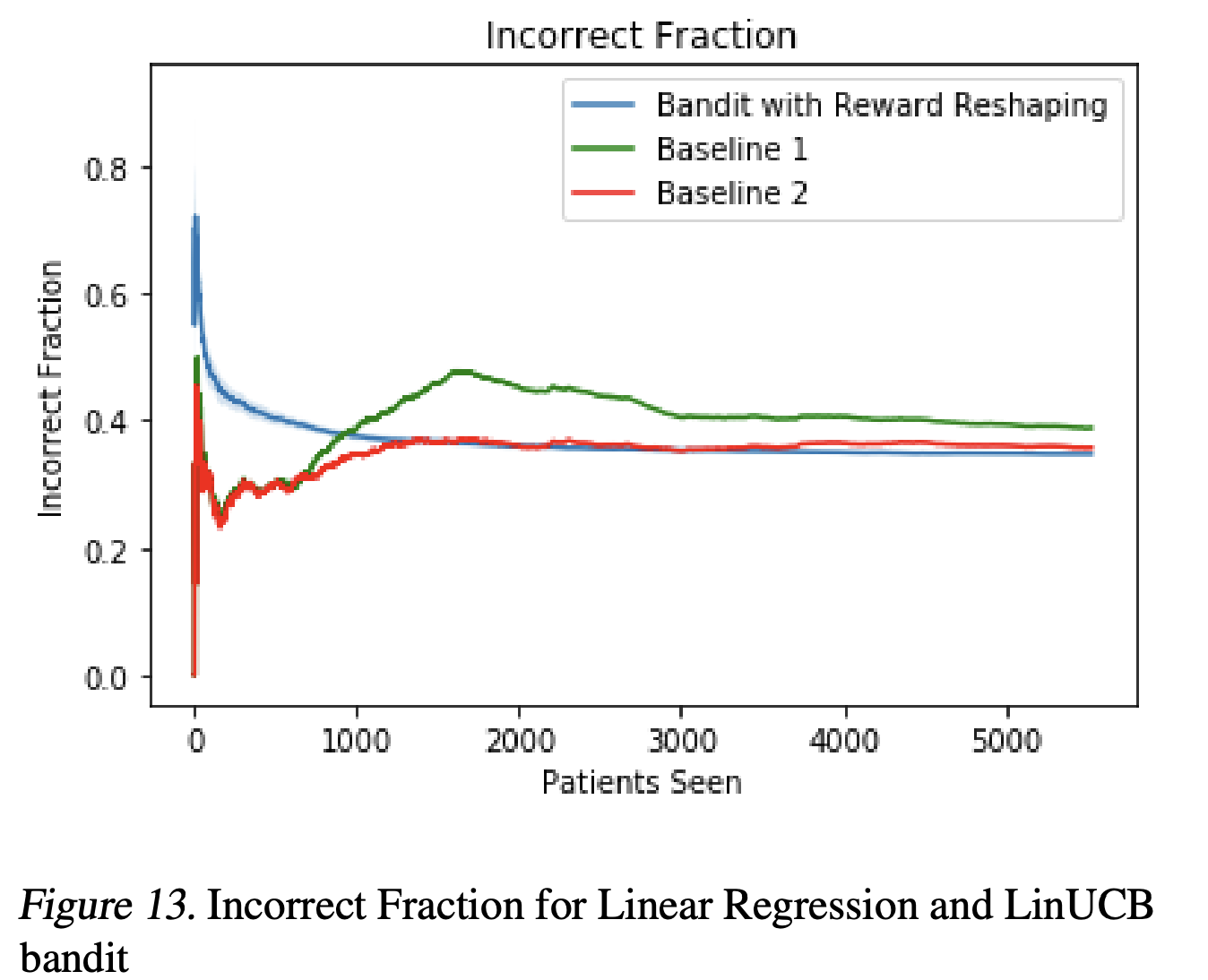}\\
\graphicspath{{/Images/Image14.png}}
\includegraphics[width = 7cm]{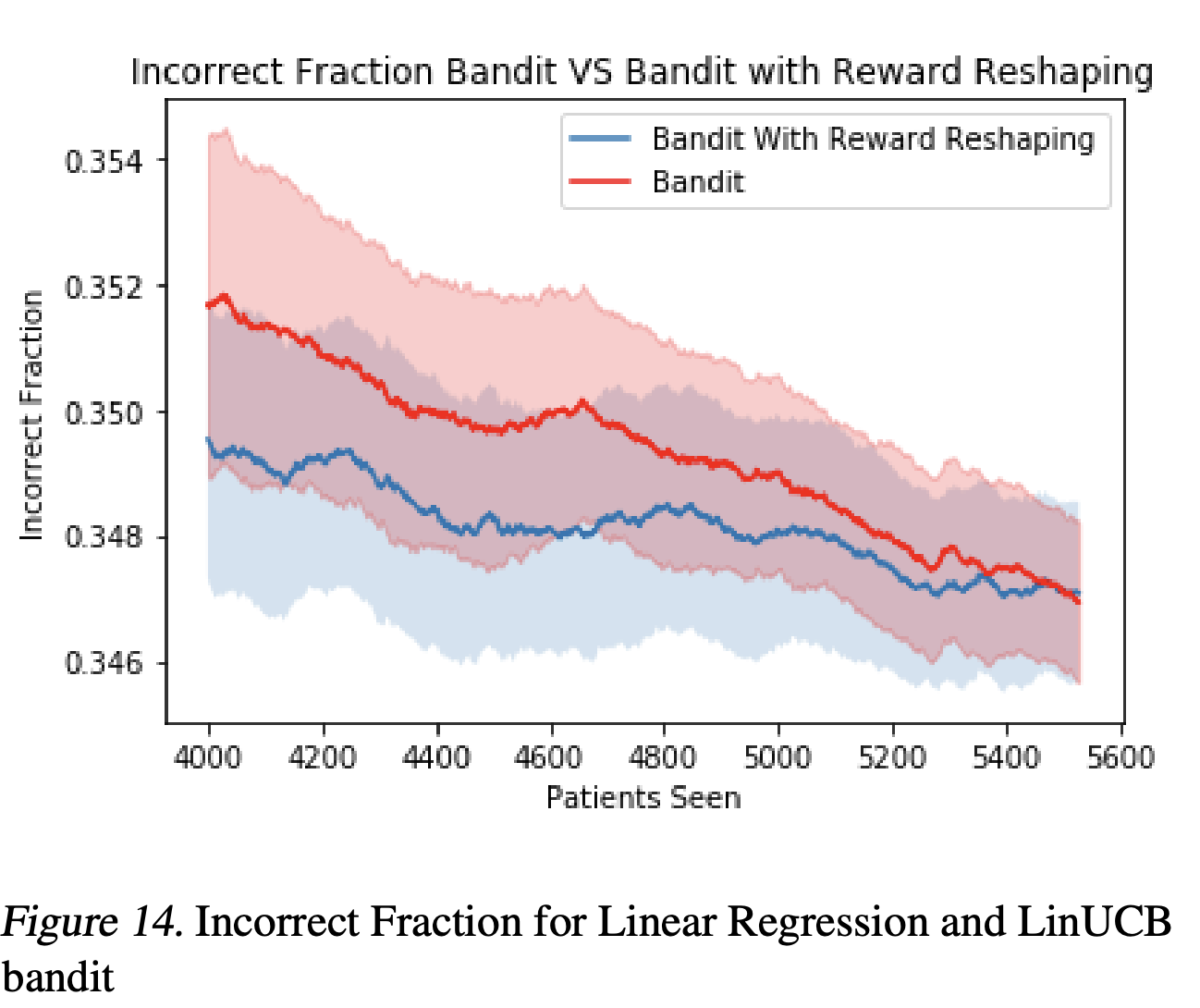}\\
Finally, to have a holistic assessment of the effectiveness of reward reshaping, we compare the performance of all the non-baseline models, LinUCB and supervised learning, with and without reward reshaping and show the result in Figure 15. From Figure 15, we can see that the models with reshaped reward benefited from a decreased incorrect fraction but were overlapping in confidence intervals with their standard reward counterparts. Figure 15 also shows the superior performance of the supervised learning model to the LinUCB bandit. Our experiments thus indicate that for problems with variable outcomes, reshaping rewards with the context of the problem in mind can lead to small performance improvements. Regarding the Warfarin dosage prediction problem,small boosts in performance are significant and substantial as it could mean the difference between a patient’s condition improving or potential adverse effects arising.\\
\graphicspath{{/Images/Image15.png}}
\includegraphics[width = 7cm]{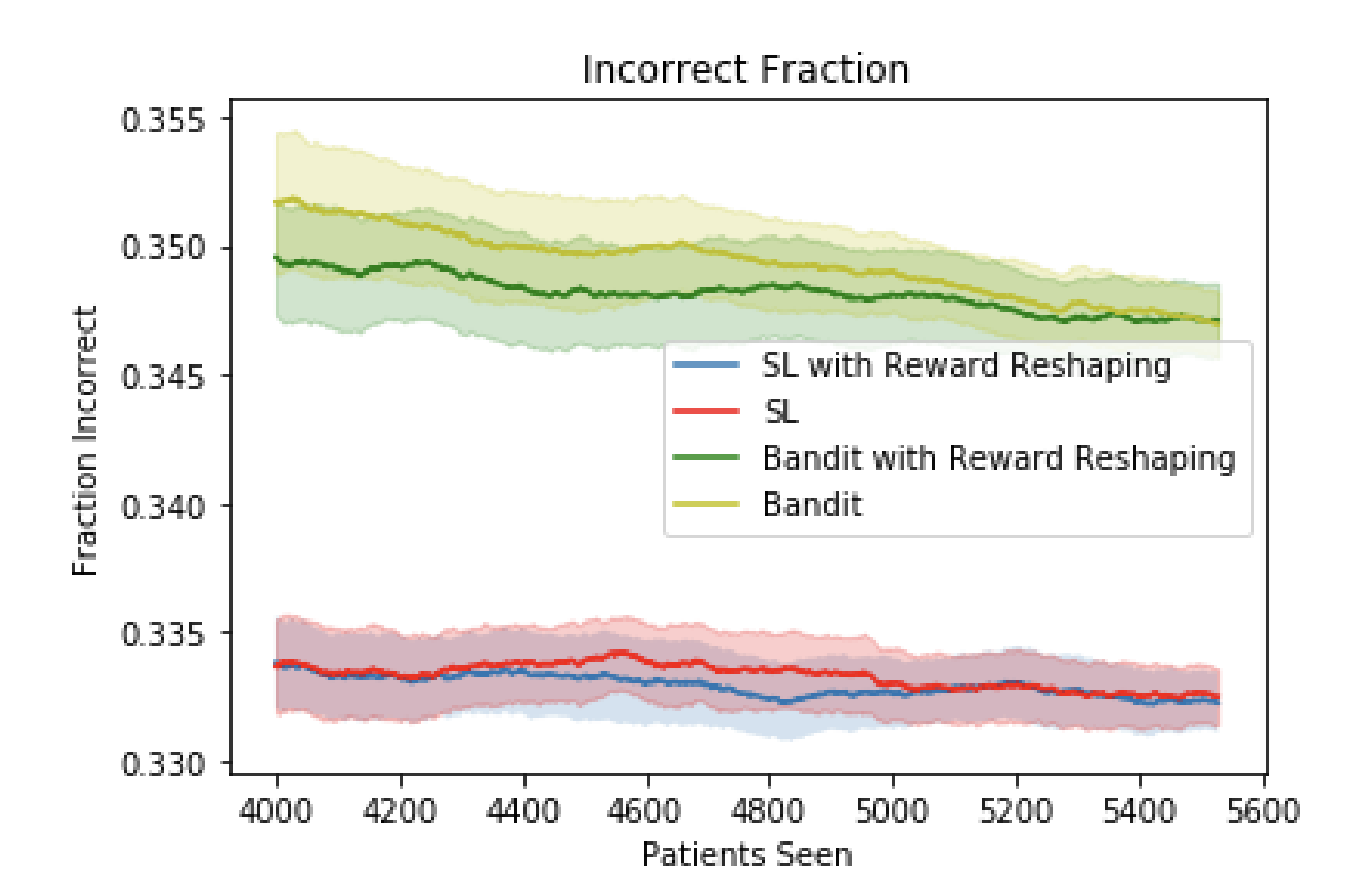}\\

\section{Conclusions} \label{Conclusions}
Following various experiments, it is clear that LinUCB bandit algorithm has a substantially better performance than the fixed dose baseline and the linear combination baseline. Linear regression on the actual dosage gives us a stronger predictor than LinUCB with the buckets, so it’s an upper bound on the performance of LinUCB. Reward shaping helps with improving accuracy with all models, though only marginally. The results from this paper make it clear that artificial intelligence and reinforcement learning are effective in improving performance on tasks such as the Warfarin dosage prediction problem and others where personalized decisions on the individual-level are beneficial and imperative to the problem. 
\subsection{Future Work}
Future work for the Warfarin dosage prediction problem could include further study with Reward Reshaping that has a stronger physiological backing to how rewards should be 
 \section{References}
 \begin{itemize}
 \item
  International Warfarin Pharmacogenetics Consortium. Estimation of the warfarin dose with clinical and pharmacogenetic data. N Engl J Med. 2009;360(8):753–764., 2009. doi: 10.1056/NEJMoa0809329.
\item
 Li, L., Chu, W., Langford, J., and Schapire, R. E. A
 contextual-bandit approach to personalized news article recommendation. Proceedings of the 19th international conference on World wide web - WWW ’10, 2010. doi: 10.1145/1772690.1772758. URL http: //dx.doi.org/10.1145/1772690.1772758.
 \end{itemize}

\end{document}